\documentclass{llncs}

\usepackage{llncsdoc}

\usepackage{graphicx}
\usepackage{caption}
\usepackage{subcaption}
\usepackage{mathtools}

\usepackage{amsmath}
\usepackage{amssymb}

\usepackage[inline]{enumitem}
\usepackage{xfrac}

\usepackage[british]{babel}
\allowhyphens

\usepackage{hyperref}
\usepackage{pbox}
\usepackage[super]{nth}
\usepackage{arydshln}

\usepackage{bm}
\usepackage{xspace}
\usepackage{mathrsfs}

\newcommand\Matrix[1]{\bm{#1}}
\newcommand\MatrixSpace[2]{\mathbb{R}^{#1\times#2}}
\newcommand\Graph[1]{\bm{\mathcal{#1}}}
\newcommand\ie{i.e.\xspace}

\newcommand\etal{et al.\xspace}
\newcommand\Reals{\mathbb{R}}

\newcommand\Integers{\mathbb{Z}}
\newcommand\sequence[3]{(#1_#2)_{#2\in#3}}
\newcommand\Id{\Matrix{I}}
\DeclareMathOperator{\relu}{ReLU}
\DeclareMathOperator{\softmax}{softmax}
\DeclareMathOperator{\LSTM}{LSTM}
\newcommand\vect[1]{{\boldsymbol{#1}}}

\newcommand\Set[1]{{\bm{\mathcal{#1}}}}
\newcommand\muacro[1]{\texttt{#1}} 
\newcommand\card[1]{{|#1|}}

\DeclareMathOperator{\graph}{GF}
\newcommand\Transpose[1]{{#1}^{\prime}}
\newcommand\Minus{\textrm{-}}

\def\vect#1{{\bm{#1}}}
\def\muacro#1{\texttt{#1}} 

\begin{document}

\title{Dynamic Graph Convolutional Networks}
\author{Franco Manessi\inst{1} \and Alessandro Rozza\inst{1} \and Mario Manzo\inst{2}}
\institute{Research Team - Waynaut\\
\email{\{name.surname\}@waynaut.com}
\and
Servizi IT - Universit\`a degli Studi di Napoli ``Parthenope''\\
\email{mario.manzo@uniparthenope.it}}
\maketitle
\begin{abstract}
Many different classification tasks need to manage structured data, which are usually modeled as graphs.
Moreover, these graphs can be dynamic, meaning that the vertices/edges of each graph may change during time.
Our goal is to jointly exploit structured data and temporal information through the use of a neural network model. To the best of our knowledge, this task has not been addressed using these kind of architectures.
For this reason, we propose two novel approaches, which combine Long Short-Term Memory networks and Graph Convolutional Networks to learn long short-term dependencies together with graph structure.
The quality of our methods is confirmed by the promising results achieved.
\end{abstract}

\section{Introduction}\label{sec:introduction}

In machine learning, data are usually described as points in a vector space ($\vect{x} \in \Reals^{d}$). Nowadays, structured data are ubiquitous and the capability to capture the structural relationships among the points can be particularly useful to improve the effectiveness of the models learned on them.

To this aim, graphs are widely employed to represent this kind of information in
terms of nodes/vertices and edges including the local and spatial information arising from data. 
Consider a $d$-dimensional dataset $\Set{X}=\{\vect{x}^{1}, \dots,\vect{x}^{n}\} \subset \Reals^{d}$, the graph is extracted from $\Set{X}$ by considering each point as a node and computing the edge weights by means of a function. We obtain a new data representation $\Graph{G}=(\Set{V},\Set{E})$, where $\Set{V}$ is a set, which contains vertices, and $\Set{E}$ is a set of weighted pairs of vertices (edges). 

Applications to a graph domain can be usually divided into two main categories, called \emph{vertex-focused} and \emph{graph-focused} applications.
For simplicity of exposition, we just consider the classification problem.\footnote{Notice that, the proposed formulation can be trivially rewritten for the regression problem.}
Under this setting, the \emph{vertex-focused} applications are characterized by a set of labels $\vect{\mathcal{L}}=\{1,\dots,k\}$, a dataset $\Set{X}=\{\vect{x}^{1},\dots,\vect{x}^{l},\vect{x}^{l+1},\dots,\vect{x}^{n}\} \subset \Reals^{d}$, the related graph $\Graph{G}$, and we assume that the first $l$ points $\vect{x}^{i}$ (where $1 \leq i \leq l$) are labeled and the remaining $\vect{x}^{u}$ (where $l+1 \leq u \leq n$) are unlabeled. The goal is to classify the unlabeled nodes exploiting the combination of their features and the graph structure by means of a semi-supervised learning approach.
Instead, \emph{graph-focused} applications are related to the goal of learning a function $f$ that maps different graphs to integer values by taking into account the features of the nodes of each graph: $f(\Graph{G}^i, \Set{X}^i) \in \Set{L}$. This task can usually be solved using a supervised classification approach on the graph structures. 

A number of research works are devoted to classify structured data both for \emph{vertex-focused} and \emph{graph-focused} applications~\cite{Jain16,rozza14,Xu13,Zhao13}.
Nevertheless, there is a major limitation in existing studies, most of these research works are focused on static graphs. However, many real-world structured data are dynamic and
nodes/edges in the graphs may change during time. In such dynamic scenario, temporal information can also play an important role.

In the last decade, (deep) neural networks have shown their great power and flexibility by learning to represent the world as a nested hierarchy of concepts, achieving outstanding results in many different fields of application. It is important to underline that, just a few research works have been devoted to encode the graph structure directly using a neural network model~\cite{Bruna13,Defferrard16,duvenaud15,kipf2016semi,Li15,Scarselli09}. Among them, to the best of our knowledge, no one is able to manage dynamic graphs.

To exploit both structured data and temporal information through the use of a neural network model, we propose two novel approaches that combine Long Short Term-Memory networks (\muacro{LSTM}s,~\cite{Hochreiter97}) and Graph Convolutional Networks (\muacro{GCN}s,~\cite{kipf2016semi}). Both of them are able to deal with \emph{vertex-focused} applications. These techniques are respectively able to capture temporal information and to properly manage structured data. Furthermore, we have also extended our approaches to deal with \emph{graph-focused} applications.

\muacro{LSTM}s are a special kind of Recurrent Neural Networks (\muacro{RNN}s,~\cite{Jain99}), 
which are able to improve the learning of long short-term dependencies. 
All \muacro{RNN}s have the form of a chain of repeating modules of neural networks. 
Precisely, \muacro{RNN}s are artificial neural networks where connections among units form a directed cycle. This creates an internal state of the network which allows it to exhibit dynamic temporal behavior.
In standard \muacro{RNN}s, the repeating module is based on a simple structure, such as a single (hyperbolic tangent) unit. \muacro{LSTM}s extend the repeating module by combining four interacting units.

\muacro{GCN} is a neural network model that directly encodes graph structure, which is trained on a supervised target loss for all the nodes with labels. 
This approach is able to distribute the gradient information from the supervised loss and to enable it to learn representations exploiting both labeled and unlabeled nodes, thus achieving state-of-the-art results. 

The paper is organized as follows: in Section \ref{sec:related} the most related methods are summarized. In Section \ref{sec:method} we describe our approaches. In Section \ref{sec:result} a comparison with baseline methodologies is presented. Section \ref{sec:conclusion} closes the paper by discussing our findings and potential future extensions.

\section{Related Work}\label{sec:related}
Many important real-world datasets are in graph form; among all, it is enough to consider: knowledge graphs, social networks, protein-interaction networks, and the World Wide Web.

To deal with this kind of data achieving good classification results, the traditional approaches proposed in literature mainly follow two different directions: to identify structural properties as features to manage them using traditional learning methods, or to propagate the labels to obtain a direct classification.

Zhu \etal~\cite{zhu03} propose a semi-supervised learning algorithm based on a Gaussian random field model (also known as Label Propagation). The learning problem is formulated as Gaussian random fields on graphs, where a field is described in terms of harmonic functions, and is efficiently solved using matrix methods or belief propagation.
Xu \etal~\cite{Xu13} present a semi-supervised factor graph model that is able to exploit the relationships among nodes. In this approach, each vertex is modeled as a variable node and the various relationships are modeled as factor nodes.
Grover and Leskovec, in~\cite{Grover16}, present an efficient and scalable algorithm for feature learning in networks that optimizes a novel network-aware, neighborhood preserving objective function using Stochastic Gradient Descent.
Perozzi \etal~\cite{Perozzi14} propose an approach called DeepWalk. This technique uses truncated random walks to efficiently learn representations for vertices in graphs. These latent
representations, which encode graph relations in a vector space, can be easily exploited by statistical models thus producing state-of-the-art results. 

Unfortunately, the described techniques are not able to deal with graphs that dynamically change in time (nodes/edges in the graphs may change during time).
There is a small amount of methodologies that have been designed to classify nodes in dynamic networks~\cite{Li13,Yao14}.
Li \etal~\cite{Li13} propose an approach that is able to learn the latent feature representation and to capture the dynamic patterns. 
Yao \etal~\cite{Yao14} present a Support Vector Machines-based approach that combines the support vectors of the previous temporal instant with the current training data to exploit temporal relationships.
Pei \etal~\cite{pei16} define an approach called dynamic Factor Graph Model for node classification in dynamic social networks. 
More precisely, this approach organizes the dynamic graph data in a sequence of graphs. Three types of factors, called node factor, correlation factor and dynamic factor, are designed to respectively capture node features, node correlations and temporal correlations.
Node factor and correlation factor are designed to capture the global and local properties of the graph structures while the dynamic factor exploits the temporal information.

It is important to underline that, very little attention has been devoted to the generalization of neural network models to structured datasets.
In the last couple of years, a number of research works have revisited the problem of generalizing neural networks to work on arbitrarily structured graphs~\cite{Bruna13,Defferrard16,duvenaud15,kipf2016semi,Li15,Scarselli09}, some of them achieving promising results in domains that have been previously dominated by other techniques.
Scarselli \etal~\cite{Scarselli09} formalize a novel neural network model, called Graph Neural Network (\muacro{GNN}s). This model is based on extending a neural network method with the purpose of
processing data in form of graph structures. The \muacro{GNN}s model can process different types of graphs (e.g., acyclic, cyclic, directed, and undirected) and it maps a graph and its nodes into a $D$-dimensional Euclidean space to learn the final classification/regression model. 
Li \etal~\cite{Li15} extend the \muacro{GNN} model, by relaxing the contractivity requirement of the propagation step through the use of Gated Recurrent Unit~\cite{cho14}, and by predicting sequence of outputs from a single input graph.
Bruna \etal~\cite{Bruna13} describe two generalizations of Convolutional Neural Networks (\muacro{CNN}s,~\cite{Goodfellow16}). 
Precisely, the authors propose two variants: one based on a hierarchical clustering of the domain and another based on the spectrum of the Laplacian graph. 
Duvenaud \etal~\cite{duvenaud15} present another variant of \muacro{CNN}s working on graph structures. This model allows an end-to-end learning on graphs of arbitrary size and shape. 
Defferrard \etal~\cite{Defferrard16} introduce a formulation of \muacro{CNN}s in the context of spectral graph theory. The model provides efficient numerical schemes to design fast
localized convolutional filters on graphs. It is important to notice that, it reaches the same computational complexity of classical \muacro{CNN}s working on any graph structure.
Kipf and Welling \cite{kipf2016semi} propose an approach for semi-supervised learning on graph-structured data (\muacro{GCN}s) based on \muacro{CNN}s. In their work, they exploit a localized first-order approximation of the spectral graph convolutions framework~\cite{Hammond11}. Their model linearly scales in the number of graph edges and learns hidden layer representations encoding local and structural graph features. 

Notice that, these neural network architectures are not able to properly deal with temporal information.

\section{Our Approaches}\label{sec:method}
In this section, we introduce two novel network architectures to deal with
\emph{vertex}/\emph{graph-focused} applications. Both of them rely on the
following intuitions:
\begin{itemize}[nolistsep]
    \item \muacro{GCN}s can effectively deal with graph-structured information,
    but they lack the ability to handle data structures that change during time.
    This limitation is (at least) twofold:
    \begin{enumerate*}[label=(\roman*)]
        \item inability to manage dynamic vertex features,
        \item inability to manage dynamic edge connections.
    \end{enumerate*}
    \item \muacro{LSTM}s excel in finding long short-term dependencies, but they
    lack the ability to explicitly exploit graph-structured information within
    it.
\end{itemize}

Due to the dynamic nature of the tasks we are interested in solving, 
the new network architectures proposed in this paper will work on ordered
sequences of graphs and ordered sequences of vertex features. Notice that, for
sequences of length one, this reduces to the 
\emph{vertex/graph-focused} applications described in
Section~\ref{sec:introduction}.

Our contributions are based on the idea of combining an extension of the Graph
Convolution (\muacro{GC}, the fundamental layer of the \muacro{GCN}s) and a
modified version of \muacro{LSTM}, thus to learn the downstream recurrent units by
exploiting both graph structured data and vertex features.

We propose two \muacro{GC}-like layers that take as input a graph
sequence and the corresponding ordered sequence of vertex features, and they
output an ordered sequence of a new vertex representation. These layers are:
\begin{itemize}[nolistsep]
    \item the \emph{Waterfall Dynamic-\muacro{GC}} layer, which performs at each
    step of the sequence a graph convolution on the vertex input sequence. An
    important feature of this layer is that the trainable parameters of each
    graph convolution are shared among the various step of the sequence;
    \item the \emph{Concatenate Dynamic-\muacro{GC}} layer, which performs at
    each step of the sequence a graph convolution on the vertex input features,
    and concatenates it to the input. Again, the trainable parameters are shared
    among the steps in the sequence.
\end{itemize}

Each of the two layers can jointly be used with a modified version of
\muacro{LSTM} to perform a semi-supervised classification of sequence of
vertices or a supervised classification of sequence of graphs. The difference
between the two tasks just consists in how we perform the last processing of the
data (for further details, see Equation~\eqref{eq:vertex-architectures} and
Equation~\eqref{eq:graph-architectures}).

In the following section we will provide the mathematical definitions of the two
modified \muacro{GC} layers, the modified version of \muacro{LSTM}, as well as
some other handy definitions that will be useful when we will describe the final
network architectures.

\subsection{Definitions}\label{sec:definitions}
Let $\sequence{\Graph{G}}{i}{\Integers_T}$ with $\Integers_T \coloneqq \{1, 2,
\ldots, T \}$ be a finite sequence of undirected graphs $\Graph{G}_i =
(\Set{V}_i, \Set{E}_i)$, with $\Set{V}_i = \Set{V}$ $\forall i \in
\Integers_T$, \ie all the graphs in the sequence share the same vertices.
Considering the graph $\Graph{G}_i$, for each vertex $v^{k} \in \Set{V}$ let
$\vect{x}^{k}_i \in \Reals^d$ be the corresponding feature vector. Each step $i$
in the sequence $\Integers_T$ can completely be defined by its graph
$\Graph{G}_i$ (modeled by the adjacency matrix\footnote{Notice that, the
adjacency matrices can be either weighted or unweighted.} $\Matrix{A}_i$) and by
the vertex-features matrix $\Matrix{X}_i \in \MatrixSpace{\card{\Set{V}}}{d}$
(the matrix whose row vectors are the $\vect{x}^{k}_i$).

We will denote with $[\Matrix{Y}]_{i,j}$ the $i$-th row, $j$-th column element
of the matrix $\Matrix{Y}$, and with $\Transpose{\Matrix{Y}}$ the transpose of
$\Matrix{Y}$. $\Matrix{I}_d$ is the identity matrix of $\Reals^d$; $\softmax$
and $\relu$ are the \emph{soft-maximum} and the \emph{rectified
linear unit} functions \cite{Goodfellow16}.

The matrix $\Matrix{P}\in\MatrixSpace{d}{d}$ is a \emph{projector} on $\Reals^d$
if it is a symmetric, positive semi-definite matrix with $\Matrix{P}^2 =
\Matrix{P}$. In particular, it is a \emph{diagonal projector} if it is a
diagonal matrix (with possibly some zero entries on the main diagonal). In other
words, a diagonal projector on $\Reals^d$ is diagonal matrix with some $1$s on
the main diagonal, that when it is right-multiplied by a $d$-dimensional column
vector $\vect{v}$ it zeroes out all the entries of $\vect{v}$ corresponding to
the zeros on the main diagonal of $\Matrix{P}$:
\begin{equation*}
    \overbracket{
        \begin{pmatrix}
            1 & 0 & 0 & 0 \\
            0 & 0 & 0 & 0 \\
            0 & 0 & 1 & 0 \\
            0 & 0 & 0 & 1 
        \end{pmatrix}
    }^{\Matrix{P}}
    \overbracket{
        \begin{pmatrix}
            a \\
            b \\
            c \\
            d 
        \end{pmatrix}
    }^{\vect{v}}
    =
    \overbracket{
        \begin{pmatrix}
            a \\
            0 \\
            c \\
            d 
        \end{pmatrix}
    }^{\Matrix{P}\vect{v}}.
\end{equation*}

We recall here the mathematics of the \muacro{GC} layer \cite{kipf2016semi} and
the \muacro{LSTM} \cite{Hochreiter97}, since they are the basic building blocks of
our contribution. Given a graph with adjacency matrix
$\Matrix{A}\in\MatrixSpace{\card{\Set{V}}}{\card{\Set{V}}}$ and vertex-feature
matrix $\Matrix{X}\in\MatrixSpace{\card{\Set{V}}}{d}$, the \muacro{GC} layer
with $M$ output nodes is defined as the function $\operatorname{GC}_M:\,
\MatrixSpace{\card{\Set{V}}}{d} \times
\MatrixSpace{\card{\Set{V}}}{\card{\Set{V}}}\to
\MatrixSpace{\card{\Set{V}}}{M}$, such that $\operatorname{GC}_M(\Matrix{X},
\Matrix{A}) \coloneqq
\relu(\hat{\Matrix{A}}\Matrix{X}\Matrix{B})$, where
$\Matrix{B}\in\MatrixSpace{d}{M}$ is a weight matrix and $\hat{\Matrix{A}}$ is
the re-normalized adjacency matrix, \ie $\hat{\Matrix{A}} \coloneqq
\tilde{\Matrix{D}}^{\,\Minus\sfrac{1}{2}}
\tilde{\Matrix{A}} \tilde{\Matrix{D}}^{\,\Minus\sfrac{1}{2}}$ with
$\tilde{\Matrix{A}} \coloneqq \Matrix{A} + \Id_{\card{\Set{V}}}$ and
$[\tilde{\Matrix{D}}]_{kk}\coloneqq \sum_l [\tilde{\Matrix{A}}]_{kl}$.

Given the sequence $\sequence{\vect{x}}{i}{\Integers_T}$ with $\vect{x}_i$
$d$-dimensional row vectors for each $i\in\Integers_T$, a \emph{returning
sequence-\muacro{LSTM}} with $N$ output nodes, is the function
$\LSTM_N:\,\sequence{\vect{x}}{i}{\Integers_T} \mapsto
\sequence{\vect{h}}{i}{\Integers_T}$, with $\vect{h}_i\in\Reals^N$ and
\begin{align*}
  & \vect{h}_i = \vect{o}_i \odot \tanh(\vect{c}_i), 
  && \vect{f}_i = \sigma(\vect{x}_i \Matrix{W}_f + \vect{h}_{i-1} \Matrix{U}_f 
    + \vect{b}_f), \\
  & \vect{c}_i = \vect{j}_i \odot \widetilde{\vect{c}}_i + \vect{f}_i \odot
    \vect{c}_{i-1}, 
  && \vect{j}_i = \sigma(\vect{x}_i \Matrix{W}_j + \vect{h}_{i-1} \Matrix{U}_j +
    \vect{b}_j), \\
  & \Matrix{o}_i = \sigma(\vect{x}_i \Matrix{W}_o + \vect{h}_{i-1} \Matrix{U}_o
    + \vect{b}_o), 
  && \widetilde{\vect{c}}_i = \sigma(\vect{x}_i \Matrix{W}_c + \vect{h}_{i-1} 
    \Matrix{U}_c + \vect{b}_c),
\end{align*} 
where $\odot$ is the Hadamard product, $\sigma(x) \coloneqq 1 / (1 +
\mathrm{e}^{\Minus x})$, $\Matrix{W}_l \in \MatrixSpace{d}{N}$, $\Matrix{U}_l
\in \MatrixSpace{N}{N}$ are weight matrices and $\vect{b}_l$ are bias vectors,
with $l\in\{o, f, j, c\}$.

\begin{definition}[\muacro{wd-GC} layer]\label{def:wd-GC}
Let $\sequence{\Matrix{A}}{i}{\Integers_T}$,
$\sequence{\Matrix{X}}{i}{\Integers_T}$ be, respectively, the sequence of
adjacency matrices and the sequence of vertex-feature matrices for the
considered graph sequence $\sequence{\Graph{G}}{i}{\Integers_T}$, with
$\Matrix{A}_i\in\MatrixSpace{\card{\Set{V}}}{\card{\Set{V}}}$ and $\Matrix{X}_i
\in \MatrixSpace{\card{\Set{V}}}{d}$ $\forall i\in\Integers_T$. The
\emph{Waterfall Dynamic-\muacro{GC}} layer with $M$ output nodes is the
function $\operatorname{wd-GC}_M$ with weight matrix $\Matrix{B} \in
\MatrixSpace{d}{M}$ defined as follows:
\begin{equation*}
    \operatorname{wd-GC}_M:\,  (\sequence{\Matrix{X}}{i}{\Integers_T},
    \sequence{\Matrix{A}}{i}{\Integers_T}) \mapsto
    (\,\relu(\hat{\Matrix{A}}_i\Matrix{X}_i\Matrix{B})\,)_{i\in\Integers_T} 
\end{equation*}
where $\relu(\hat{\Matrix{A}}_i\Matrix{X}_i\Matrix{B}) \in \Reals^{\card{\Set{V}}
\times M}$, and all the $\hat{\Matrix{A}}_i$ are the re-normalized adjacency matrices of the
graph sequence $\sequence{\Graph{G}}{i}{\Integers_T}$.
\end{definition}

The \muacro{wd-GC} layer can be seen as multiple copies of a standard
\muacro{GC} layer, all of them sharing the same training weights. Then, the
resulting training parameters are $d \cdot M$, independently of the length of
the sequence.

In order to introduce the Concatenate Dynamic-\muacro{GC} layer, we recall the
definition of the \emph{Graph of a Function}: considering a function $f$ from $A$ to $B$,
$[\graph f]:\, A \to A \times B$, $x \mapsto [\graph f](x) \coloneqq (x, f(x))$.
Namely, the $\graph$ operator transforms $f$ into a function returning the concatenation between $x$ and $f(x)$.

\begin{definition}[\muacro{cd-GC} layer]\label{def:cd-GC}
Let $\sequence{\Matrix{A}}{i}{\Integers_T}$,
$\sequence{\Matrix{X}}{i}{\Integers_T}$ be, respectively, the sequence of
adjacency matrices and the sequence of vertex-feature matrices for the
considered graph sequence $\sequence{\Graph{G}}{i}{\Integers_T}$, with
$\Matrix{A}_i\in\MatrixSpace{\card{\Set{V}}}{\card{\Set{V}}}$ and $\Matrix{X}_i
\in \MatrixSpace{\card{\Set{V}}}{d}$ $\forall i\in\Integers_T$. A
\emph{Concatenate Dynamic-\muacro{GC}} layer with $M$ output nodes is the
function $\operatorname{cd-GC}_M$ with weight matrix $\Matrix{B} \in
\MatrixSpace{d}{M}$ defined as follows:
\begin{equation*}
    \operatorname{cd-GC}_M:\, (\sequence{\Matrix{X}}{i}{\Integers_T},
    \sequence{\Matrix{A}}{i}{\Integers_T}) \mapsto (\,
    [\graph\relu](\hat{\Matrix{A}}_i\Matrix{X}_i\Matrix{B})
    \,)_{i\in\Integers_T}
\end{equation*}
where $[\graph\relu](\hat{\Matrix{A}}_i\Matrix{X}_i\Matrix{B}) \in
\Reals^{\card{\Set{V}} \times (M + d)}$, and all the $\hat{\Matrix{A}}_i$ are the
re-normalized adjacency matrices of the graph sequence
$\sequence{\Graph{G}}{i}{\Integers_T}$.
\end{definition}
Intuitively, \muacro{cd-GC} is a layer made of $T$ copies of \muacro{GC}
layers, each copy acting on a specific instant of the sequence.
Each output of the $T$ copies is then concatenated with its input, thus resulting in a sequence of
graph-convoluted features together with the vertex-features matrix. Note that, the
weights $\Matrix{B}$ are shared among the $T$ copies. The number of learnable
parameters of this layer is $d \cdot (d+M)$, independently of the number of
steps in the sequence $\sequence{\Graph{G}}{i}{\Integers_T}$.

Notice that, both the input and the output of \muacro{wd-GC} and \muacro{cd-GC}
are sequences of matrices (loosely speaking, third order tensors).

We will now define three additional layers. These will help us in reducing the
clutter with the notation when we will introduce in
Section~\ref{sec:semi-supervised-vertex-sequence-classification} and
Section~\ref{sec:supervised-graph-sequence-classification} the network architectures
we have used to solve the semi-supervised classification of sequence of vertices and
the supervised classification of sequence of graphs. Precisely, they are:
\begin{enumerate*}[label=(\roman*)]
    \item the recurrent layer used to process in a parallel fashion the
    convoluted vertex features,
    \item the two final layers (one per task) used to map the previous layers
    outputs into $k$-class probability vectors.
\end{enumerate*}

\begin{definition}[\muacro{v-LSTM} layer]
    Consider $\sequence{\Matrix{Z}}{i}{\Integers_T}$ with
    $\Matrix{Z}\in\MatrixSpace{L}{M}$, the \emph{Vertex \muacro{LSTM}} layer with $N$
    output nodes is given by the function $\operatorname{v-LSTM}_N$:
    \begin{equation*}
        \operatorname{v-LSTM}_N:\, \sequence{\Matrix{Z}}{i}{\Integers_T} \mapsto
        \begin{pmatrix}
            \LSTM_N(\sequence{\Transpose{\Matrix{V}}_{1}\Matrix{Z}}{i}{\Integers_T}) \\
            \vdots\\
            \LSTM_N(\sequence{\Transpose{\Matrix{V}}_{L}\Matrix{Z}}{i}{\Integers_T})
        \end{pmatrix} \in \Reals^{L \times N \times T},
    \end{equation*}
    where $\Matrix{V}_{p}$ is the isometric embedding of $\Reals$ into
    $\Reals^L$ defined as $[\Matrix{V}_{p}]_{i,j} = \delta_{ip}$, and $\delta$
    is the Kronecker delta function. The training weights are shared among the
    $L$ copies of the \muacro{LSTM}s.
\end{definition}

\begin{definition}[\muacro{vs-FC} layer]
    Consider $\sequence{\Matrix{Z}}{i}{\Integers_T}$ with
    $\Matrix{Z}\in\MatrixSpace{L}{N}$, the \emph{Vertex Sequential Fully
    Connected} layer with $k$ output nodes is given by the function
    $\operatorname{vs-FC}_k$, parameterized by the weight matrix
    $\Matrix{W}\in\MatrixSpace{N}{k}$ and the bias matrix $\MatrixSpace{L}{k}
    \ni \Matrix{B} \coloneqq \Transpose{(\Transpose{\vect{b}}, \ldots,
    \Transpose{\vect{b}})}$:
    \begin{equation*}
        \operatorname{vs-FC}_k:\, \sequence{\Matrix{Z}}{i}{\Integers_T} \mapsto
        (\,
        \softmax(\Matrix{W}\Matrix{Z}_i + \Matrix{B})\,)_{i\in\Integers_T}
    \end{equation*}
    with $\softmax(\Matrix{W}\Matrix{Z}_i + \Matrix{B}) \in \MatrixSpace{L}{k}$.
\end{definition}

\begin{definition}[\muacro{gs-FC} layer]
    Consider $\sequence{\Matrix{Z}}{i}{\Integers_T}$ with
    $\Matrix{Z}\in\MatrixSpace{L}{N}$, the \emph{Graph Sequential Fully
    Connected} layer with $k$ output nodes is given by the function
    $\operatorname{gs-FC}_k$, parameterized by the weight matrices
    $\Matrix{W}_1\in\MatrixSpace{N}{k}$, $\Matrix{W}_2\in\MatrixSpace{1}{L}$ and
    the bias matrices $\MatrixSpace{L}{k} \ni \Matrix{B}_1 \coloneqq
    \Transpose{(\Transpose{\vect{b}}, \ldots, \Transpose{\vect{b}})}$ and 
    $\Matrix{B}_2 \in \MatrixSpace{1}{k}$:
    \begin{equation*}
        \operatorname{gs-FC}_K:\, \sequence{\Matrix{Z}}{i}{\Integers_T} \mapsto
        (\, \softmax( \Matrix{W}_2 \relu( \Matrix{W}_1 \Matrix{Z}_i +
        \Matrix{B}_1 ) + \Matrix{B}_2 )\,)_{i\in\Integers_T}
    \end{equation*}
    with $\softmax(\Matrix{W}_2\relu(\Matrix{W}_1\Matrix{Z}_i + \Matrix{B}_1) +
    \Matrix{B}_2) \in \MatrixSpace{1}{k}$.
\end{definition}

Informally:
\begin{enumerate*}[label=(\roman*)]
    \item the \muacro{v-LSTM} layer acts as $L$ copies of \muacro{LSTM}, each one evaluating
    the sequence of one row of the input tensor
    $\sequence{\Matrix{Z}}{i}{\Integers_T}$;
    \item the \muacro{vs-FC} layer acts as $T$ copies of a Fully Connected layer
    (\muacro{FC}, \cite{Goodfellow16}) with softmax activation, all the copies sharing the
    parameters. The \muacro{vs-FC} layer outputs $L$ $k$-class probability vectors for
    each step in the input sequence;
    \item the \muacro{gs-FC} layer acts as $T$ copies of two \muacro{FC} layers
    with softmax-ReLU activation, all the copies sharing the parameters. This
    layer outputs one $k$-class probability vector for each step in the input
    sequence.
\end{enumerate*}
Note that, both the input and the output of \muacro{vs-FC} and \muacro{v-LSTM}
are sequences of matrices, while for \muacro{gs-FC} the input is a sequence of
matrices and the output is a sequence of vectors.

We have now all the elements to describe our network architectures to address
both semi-supervised classification of sequence of vertices and supervised
classification of sequence of graphs.

\begin{figure*}[!t]
    \begin{subfigure}[t]{\textwidth}
        \centering
        \hspace{2mm}\includegraphics[height=2.9cm]{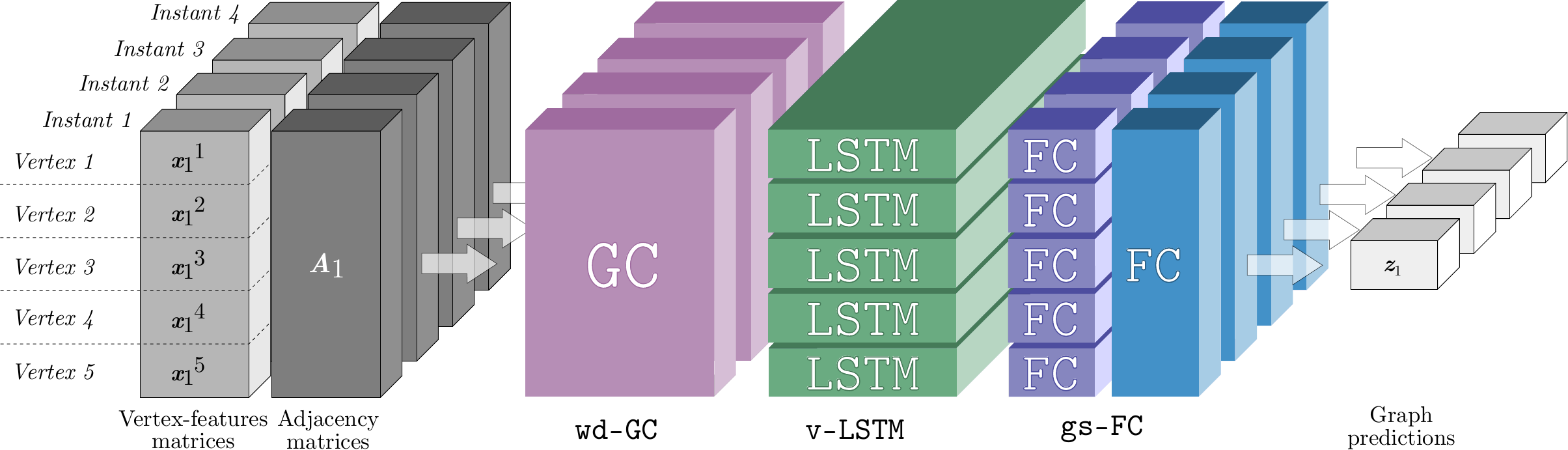}
        \caption{\label{fig:wd-GC}\muacro{WD-GCN} for classification of sequence
        of graphs.}
    \end{subfigure}

    \vspace{.5em}

    \begin{subfigure}[t]{\textwidth}
        \centering
        \includegraphics[height=2.9cm]{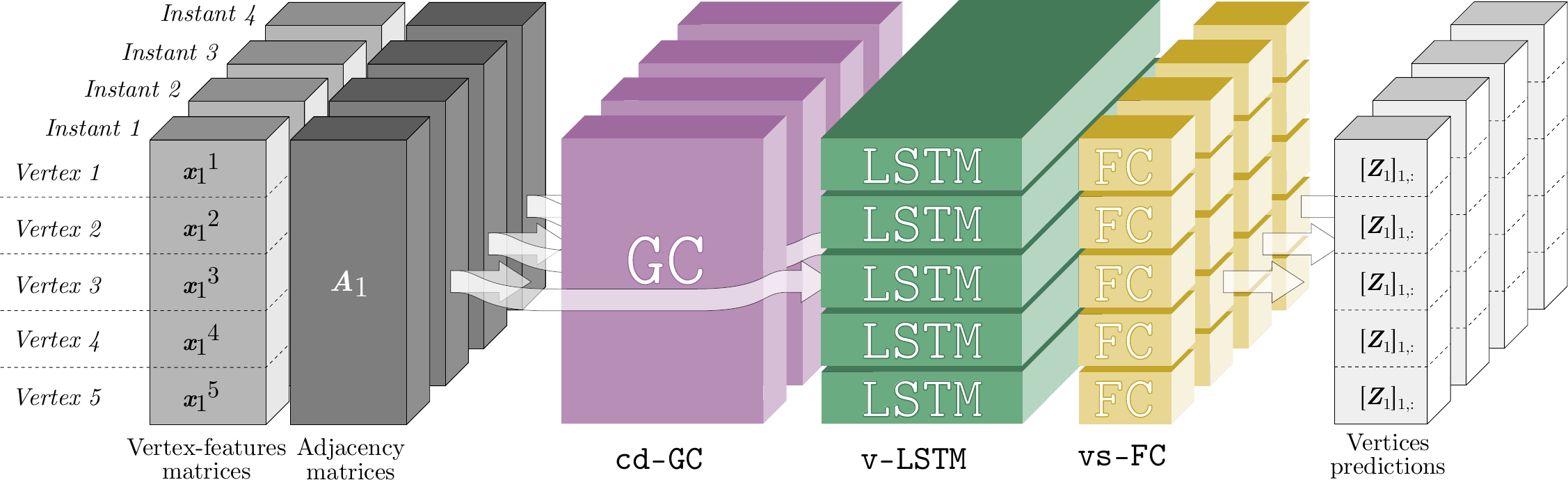}
        \caption{\label{fig:cd-GC}\muacro{CD-GCN} for classification of sequence
        of vertices.}
    \end{subfigure}
    \caption{\label{fig:two-layers}
        The figure shows two of the four network architectures presented in
        Sections~\ref{sec:semi-supervised-vertex-sequence-classification}~and~\ref{sec:supervised-graph-sequence-classification},
        both of them working on sequences of four graphs composed of five
        vertices, \ie $\sequence{\Graph{G}}{i}{\Integers_4}$, $\card{\Set{V}} =
        5$.
        (a) The \muacro{wd-GC} layer acts as four copies of a regular
        \muacro{GC} layer, each one working on an instant of the sequence. The
        output of this first layer is processed by the \muacro{v-LSTM} layer
        that acts as five copies of the \textit{returning
        sequence}-\muacro{LSTM} layer, each one working on a vertex of the
        graphs. The final \muacro{gs-FC} layer, which produces the $k$-class
        probability vector for each instant of the sequence, can be seen as the
        composition of two layers: the first one working on each vertex for
        every instant, and the following one working on \emph{all} the vertices
        at a specific instant.
        (b) The \muacro{cd-GC} and the \muacro{v-LSTM} layers work as the
        \muacro{wd-GC} and the \muacro{v-LSTM} of the Figure 1a, the only
        difference is that \muacro{v-LSTM} works both on graph convolutional
        features, as well as plain vertex features, due to the fact that
        \muacro{cd-GC} produces their concatenation. The last layer, which
        produces the $k$-class probability vector for each vertex and for each
        instant of the sequence, can be seen as $5 \times 4$ copies of a
        \muacro{FC} layer.
        }
        \vspace{-1.8em}
\end{figure*}

\subsection{Semi-Supervised Classification of Sequence of Vertices}
\label{sec:semi-supervised-vertex-sequence-classification}
\begin{definition}[Semi-Supervised Classification of Sequence of Vertices]
\label{def:semi-supervised-vertex-sequence-classification}
Let $\sequence{\Graph{G}}{i}{\Integers_T}$ be a sequence of $T$ graphs
each one made of $\card{\Set{V}}$ vertices, and
$\sequence{\Matrix{X}}{i}{\Integers_T}$ the related sequence of vertex-features
matrices.
 
Let $\sequence{\Matrix{P}^{\textrm{Lab}}}{i}{\Integers_T}$ be a sequence of
diagonal projectors on the vector space $\Reals^{\card{\Set{V}}}$. Define the
sequence $\sequence{\Matrix{P}^{\textrm{Unlab}}}{i}{\Integers_T}$ by means of
$\Matrix{P}^{\textrm{Unlab}}_i \coloneqq \Id_\card{\Set{V}} -
\Matrix{P}^{\textrm{Lab}}_i$, $\forall i \in \Integers_T$; \ie
$\Matrix{P}^{\textrm{Lab}}_i$ and $\Matrix{P}^{\textrm{Unlab}}_i$ identify the
labeled and unlabeled vertices of $\Graph{G}_i$, respectively.
Moreover, let $\sequence{\Matrix{Y}}{i}{\Integers_T}$ be a sequence of $T$
matrices with $\card{\Set{V}}$ rows and $k$ columns, satisfying the property
$\Matrix{P}^{\textrm{Lab}}_i \Matrix{Y}_i = \Matrix{Y}_i$, where the $j$-th row
of the $i$-th matrix represents the one-hot encoding of the $k$-class label of
the $j$-th vertex of the $i$-th graph in the sequence, with the $j$-th vertex
being a labeled one.
Then, \emph{semi-supervised classification of sequence of vertices} consists in
learning a function $f$ such that $\Matrix{P}^{\textrm{Lab}}_j f(\,
\sequence{\Graph{G}}{i}{\Integers_T},
\sequence{\Matrix{X}}{i}{\Integers_T} \,)_j = \Matrix{Y}_j$ and
$\Matrix{P}^{\textrm{Unlab}}_j f(\,
\sequence{\Graph{G}}{i}{\Integers_T},\allowbreak
\sequence{\Matrix{X}}{i}{\Integers_T} \,)_j$ is the right labeling for the
unlabeled vertices for each $j\in\Integers_T$.
\end{definition}

To address the above task, we propose the networks defined by the following 
functions:
\begin{subequations}\label{eq:vertex-architectures}
    \begin{align}
        \label{eq:wd-GC-vertex-architecture}
        \operatorname{v\_wd-GC\_LSTM}_{M,N,k}:\quad& \operatorname{vs-FC}_k 
            \circ \operatorname{v-LSTM}_N \circ \operatorname{wd-GC}_M, \\
        \label{eq:cd-GC-vertex-architecture}
        \operatorname{v\_cd-GC\_LSTM}_{M,N,k}:\quad& \operatorname{vs-FC}_k 
            \circ \operatorname{v-LSTM}_N \circ \operatorname{cd-GC}_M,
\end{align}
\end{subequations}
where $\circ$ denote the function composition. Both the architectures take
$(\sequence{\Matrix{X}}{i}{\Integers_T},\allowbreak
\sequence{\Matrix{A}}{i}{\Integers_T})$ as input, and produce a sequence of
matrices whose row vectors are the probabilities of each vertex of the graph:
$\sequence{\Matrix{Z}}{i}{\Integers_T}$ with $\Matrix{Z}_i \in
\MatrixSpace{\card{\Set{V}}}{k}$. For the sake of clarity, in the rest of the
paper, we will refer to the networks defined by
Equation~\eqref{eq:wd-GC-vertex-architecture} and
Equation~\eqref{eq:cd-GC-vertex-architecture} as \emph{Waterfall
Dynamic-\muacro{GCN}}(\muacro{WD-GCN}) and
\emph{Concatenate Dynamic-\muacro{GCN}} (\muacro{CD-GCN}, see Figure~\ref{fig:cd-GC}), respectively.

Since all the functions involved in the composition are differentiable, the
weights of the architectures can be learned using gradient descent methods,
employing as loss function the \emph{cross-entropy} evaluated only on the
labeled vertices:
\begin{equation*}
    \mathcal{L} = - \sum_{t\in\Integers_T} \sum_{c\in\Integers_k}
    \sum_{v\in\Integers_{\card{\Set{V}}}} [\Matrix{Y}_t]_{v,c} \log
    [\Matrix{P}^{\textrm{Lab}}_t\Matrix{Z}_t]_{v,c},
\end{equation*}
with the convention that $0\cdot\log 0 = 0$.

\subsection{Supervised Classification of Sequence of Graphs}
\label{sec:supervised-graph-sequence-classification}
\begin{definition}[Supervised Classification of Sequence of Graphs]
Let $\sequence{\Graph{G}}{i}{\Integers_T}$ be a sequence of $T$ graphs each one
made of $\card{\Set{V}}$ vertices, and $\sequence{\Matrix{X}}{i}{\Integers_T}$
the related sequence of vertex-features matrices. Moreover, let
$\sequence{\vect{y}}{i}{\Integers_T}$ be a sequence of $T$ one-hot encoded
$k$-class labels, \ie $\vect{y}_i \in \{0, 1\}^k$.
Then, \emph{graph-sequence classification task} consists in learning a
predictive function $f$ such that $f(\,\sequence{\Graph{G}}{i}{\Integers_T},
\sequence{\Matrix{X}}{i}{\Integers_T}\,) = \sequence{\vect{y}}{i}{\Integers_T}$.
\end{definition}

The proposed architectures are defined by the following functions:
\begin{subequations}\label{eq:graph-architectures}
    \begin{align}
        \label{eq:wd-GC-graph-architecture}
        \operatorname{g\_wd-GC\_LSTM}_{M,N,k}:\quad& \operatorname{gs-FC}_k 
            \circ \operatorname{v-LSTM}_N \circ \operatorname{wd-GC}_M, \\
        \label{eq:cd-GC-graph-architecture}
        \operatorname{g\_cd-GC\_LSTM}_{M,N,k}:\quad& \operatorname{gs-FC}_k 
            \circ \operatorname{v-LSTM}_N \circ \operatorname{cd-GC}_M,
    \end{align}
\end{subequations}
The two architectures take as input
$(\sequence{\Matrix{X}}{i}{\Integers_T},
\allowbreak \sequence{\Matrix{A}}{i}{\Integers_T})$. The output of
\muacro{wd-GC} and \muacro{cd-GC} is processed by a \muacro{v-LSTM}, resulting
in a $\card{\Set{V}}\times N$ matrix for each step in the sequence. It is a
\muacro{gs-FC} duty to transform this vertex-based prediction into a graph based
prediction, \ie to output a sequence of $k$-class probability vectors
$\sequence{\vect{z}}{i}{\Integers_T}$. Again, we will use \muacro{WD-GCN} (see Figure~\ref{fig:wd-GC}) and
\muacro{CD-GCN} to refer to the networks defined by
Equation~\eqref{eq:wd-GC-graph-architecture} and
Equation~\eqref{eq:cd-GC-graph-architecture}, respectively.

Also under this setting the training can be performed by means of gradient
descent methods, with the cross entropy as loss function:
\begin{equation*}
    \mathcal{L} = - \sum_{t\in\Integers_T} \sum_{c\in\Integers_k} [\vect{y}_t]_c
    \log [\vect{z}_t]_c,
\end{equation*}
with the convention $0\cdot\log 0 = 0$.

\section{Experimental Results}\label{sec:result}
In this section we describe the employed datasets, the experimental settings, and the results achieved by our approaches compared with those obtained by baseline methods.

\subsection{Datasets}
We now present the used datasets. The first one is employed to evaluate our approaches in the context of the \emph{vertex-focused} applications; instead, the second dataset is used to assess our architectures in the context of the \textit{graph-focused} applications.

Our first set of data is a subset of
\muacro{DBLP}\footnote{\url{http://dblp.uni-trier.de/xml/}}
dataset described in~\cite{pei16}. Conferences from six research communities, including
artificial intelligence and machine learning, algorithm and theory, database,
data mining, computer vision, and information retrieval, have been considered.
Precisely, the co-author relationships from $2001$ to $2010$ are considered and
data of each year is organized in a graph form. Each author represents a node in
the network and an edge between two nodes exists if two authors have
collaborated on a paper in the considered year. Note that, the resulting
adjacency matrix is \emph{unweighted}.

The node features are extracted from each temporal instant using DeepWalk
\cite{Perozzi14} and are composed of 64 values. Furthermore, we have augmented
the node features by adding the number of articles published by the authors
in each of the six communities, obtaining a features vector composed of $70$
values. This specific task belongs to the \emph{vertex-focused} applications.

The original dataset is made of $25.215$ authors across the ten
years under analysis. Each year $4.786$ authors appear on average, and $114$
authors appear all the years, with an average of $1.594$ authors appearing on
two consecutive years.

We have considered the $500$ authors with the highest number of connections during
the analyzed $10$ years, \ie the $500$ vertices among the total $25.215$ with the
highest $\sum_{t\in\Integers_{10}}\sum_i [\Matrix{A}_t]_{i,j}$, with
$\Matrix{A}_t$ the adjacency matrix at the $t$-th year. If one of the $500$
selected authors does not appear in the $t$-th year, its feature vector is set to zero.

The final dataset is composed of $10$ vertex-features matrices in
$\MatrixSpace{500}{70}$, $10$ adjacency matrices belonging to
$\MatrixSpace{500}{500}$, and each vertex belongs to one of the $6$ classes.

\muacro{CAD-120}\footnote{\url{http://pr.cs.cornell.edu/humanactivities/data.php}}
is a dataset composed of $122$ \muacro{RGB-D} videos corresponding to $10$
high-level human activities~\cite{Koppula13}.
Each video is annotated with sub-activity labels, object affordance labels,
tracked human skeleton joints and tracked object bounding boxes. The $10$ sub-activity labels are: \emph{reaching}, \emph{moving}, \emph{pouring},
\emph{eating}, \emph{drinking}, \emph{opening}, \emph{placing}, \emph{closing},
\emph{scrubbing}, \emph{null}.
Our second dataset is composed of all the data related to the detection of
sub-activities, \ie no object affordance data have been considered. Notice that, detecting the
sub-activities is a challenging problem as it involves complex interactions,
since humans can interact with multiple objects during a single activity. This
specific task belongs to the \textit{graph-focused} applications.

Each one of the $10$ high-level activities is characterized by one person, whose
$15$ joints are tracked (in position and orientation) in the $3$D space for
each frame of the sequence. Moreover, in each high-level activity appears a
variable number of objects, for which are registered their bounding boxes in the
video frame together with the transformation matrix matching extracted SIFT
features \cite{lowe1999object} from the frame to the ones of the previous frame. Furthermore, there are $19$ objects involved in the videos.

We have built a graph for each video frame: the vertices are the $15$ skeleton joints
plus the $19$ objects, while the \emph{weighted} adjacency matrix has been derived
by employing Euclidean distance. Precisely, among two skeleton joints the edge weight is given by the Euclidean distance
    between their $3$D positions; among two objects it is the $2$D distance between the centroids of
    their bounding boxes; among an object and a skeleton joint it is the $2$D distance between
    the centroid of the object bounding box and the skeleton joint projection into the
    $2$D video frame.
All the distances have been scaled between zero and one. 
When an object does not appear in a frame, its related row and column in the adjacency matrix is set to zero.

Since the videos have different lengths, we have padded all the sequences to match the longest one, which has $1.298$ frames.

Finally, the feature columns have been standardized. 
The resulting dataset is composed of $122 \times 1.298$ vertex-feature matrices belonging to $\MatrixSpace{34}{24}$, $122 \times 1.298$ adjacency matrices (in $\MatrixSpace{34}{34}$), and each graph belongs to one of the $10$ classes.

\subsection{Experimental Settings}\label{sec:experimental-settings}
In our experiments, we have compared the results achieved by the proposed architectures with those obtained by other
baseline networks (see Section~\ref{sec:results} for a full description of the chosen baselines).

For the baselines that are not able to explicitly exploit sequentiality in the data, we have flatten
the temporal dimension of all the sequences, thus considering the same point in two different time instants as two different training samples.

The hyper-parameters of all the networks (in terms of number of nodes of each
layer and dropout rate) have been appropriately tuned by means of a grid
approach. The performances are assessed employing $10$ iterations of Monte Carlo Cross-Validation\footnote{This approach randomly selects (without replacement) some fraction of the data to build the training set, and it assignes the rest of the samples to the test set. This process is repeated multiple times, generating (at random) new training and test partitions each time. Notice that, in our experiments, the training set is further split
into training and validation.} preserving the percentage of samples for each class.
It is important to underline that, the $10$ train/test sets are generated once, and they are used to evaluate all the architectures, to keep the experiments as fair as possible.
To assess the performances of all the considered
architectures we have employed \emph{Accuracy} and \emph{Unweighted F1 Measure}\footnote{The Unweighted F1 Measure evaluates the F1 scores for each label class, and find their unweighted mean: 
$\frac{1}{k} \sum_{c\in\Integers_k} \frac{2p_c r_c}{p_c + r_c}$,
 where $p_c$ and $r_c$ are the \emph{precision} and the \emph{recall} of the class $c$.}.
Moreover, the training phase has been performed using Adam \cite{kingma2014adam}
for a maximum of $100$ epochs, and for each network (independently for Accuracy
and F1 Measure) we have selected the epoch where the learned model achieved the
best performance on the validation set using the learned model to finally
assess the performance on the test set.

\subsection{Results}\label{sec:results}

\subsubsection{DBLP}
We have compared the approaches proposed in Section~\ref{sec:semi-supervised-vertex-sequence-classification} ($\muacro{WD-GCN}$ and $\muacro{CD-GCN}$) against the following baseline methodologies:
\begin{enumerate*}[label=(\roman*)]
    \item a \muacro{GCN} composed of two layers,
    \item a network made of two \muacro{FC} layers,
    \item a network composed of \muacro{LSTM}+\muacro{FC},
    \item and a deeper architecture made of \muacro{FC}+\muacro{LSTM}+\muacro{FC}.
\end{enumerate*}
Note that, the \muacro{FC} is a Fully Connected layer; when it appears as the first layer of a network it employes a $\relu$ activation, instead a $\softmax$ activation is used when it is the last layer of a network.

The test set contains $30\%$ of the $500$ vertices. Moreover, $20\%$ of the remaining vertices (the training ones) have been used for validation purposes. It is important to underline that, an unlabeled vertex remains unlabeled for all the years in the sequence, \ie considering Definition~\ref{def:semi-supervised-vertex-sequence-classification}, $\Matrix{P}^{\textrm{Lab}}_i = \Matrix{P}^{\textrm{Lab}}$, $\forall i \in \Integers_T$.

In Table~\ref{tab:DBLP-results-grid}, the best hyper-parameter configurations together with the test results of all the evaluated architectures are presented.
\begin{table}[t]
    \caption{\label{tab:DBLP-results-grid}Results of the evaluated architectures on semi-supervised classification of sequence of vertices employing the \muacro{DBLP}
    dataset. We have tested the statistical significance of our result by means of \emph{Wilcoxon test}, obtaining a p-value $<0.6\%$ when we have compared \muacro{WD-GCN} and \muacro{CD-GCN} against all the baselines for both the employed scores.
    }
\smallskip
\centering
\resizebox{.9\textwidth}{!}{
\begin{tabular}{lllllcllc}
\cline{5-6}
\cline{8-9}
\noalign{\smallskip}
& & & & \multicolumn{2}{c}{Accuracy} & & \multicolumn{2}{c}{Unweighted F1 Measure} \\
\hline\noalign{\smallskip}
Network & Hyper-params & Grid & \hspace{1em} & \pbox{5cm}{Best \\ Config.} & \pbox{5cm}{Performance\\mean $\pm$ std} & \hspace{1em} & \pbox{5cm}{Best \\ Config.} & \pbox{5cm}{Performance\\mean $\pm$ std} \\
\noalign{\smallskip}
\hline
\noalign{\smallskip}
\muacro{FC}+\muacro{FC} & \pbox{5cm}{\nth{1} \muacro{FC} nodes: \\ dropout:} & 
\pbox{5cm}{$\{150, 200, 250, 300, 350, 400\}$ \\ $\{0\%, 10\%, 20\%, 30\%, 40\%, 50\% \}$} & & \pbox{5cm}{250 \\ 50\%} & $49.1\% \pm 1.2\%$ & & \pbox{5cm}{250 \\ 40\%} & $48.2\% \pm 1.3\%$ \\
\noalign{\smallskip}
\hdashline[1pt/3pt]
\noalign{\smallskip}
\muacro{GC}+\muacro{GC} & \pbox{5cm}{\nth{1} \muacro{GC} nodes: \\ dropout:} &
\pbox{5cm}{$\{150, 200, 250, 300, 350, 400\}$ \\ $\{0\%, 10\%, 20\%, 30\%, 40\%, 50\% \}$} & & \pbox{5cm}{350 \\ 50\%} & $54.8\% \pm 1.4\%$ & & \pbox{5cm}{350 \\ 10\%} & $54.7\% \pm 1.7\%$ \\
\noalign{\smallskip}
\hdashline[1pt/3pt]
\noalign{\smallskip}
\muacro{LSTM}+\muacro{FC} & \pbox{5cm}{\muacro{LSTM} nodes: \\ dropout:} & 
\pbox{5cm}{$\{100, 150, 200, 300, 400\}$ \\ $\{0\%, 10\%, 20\%, 30\%, 40\%, 50\% \}$}& & \pbox{5cm}{100 \\ 0\%} & $60.1\% \pm 2.1\%$ & & \pbox{5cm}{100 \\ 0\%} & $60.4\% \pm 2.3\%$ \\
\noalign{\smallskip}
\hdashline[1pt/3pt]
\noalign{\smallskip}
\muacro{FC}+\muacro{LSTM}+\muacro{FC} & \pbox{5cm}{\muacro{FC} nodes: \\ \muacro{LSTM} nodes: \\ dropout:} & 
\pbox{5cm}{$\{100, 200, 300, 400\}$ \\ $\{100, 200, 300, 400\}$ \\ $\{0\%, 10\%, 20\%, 30\%, 40\%, 50\% \}$}& & \pbox{5cm}{300 \\ 300 \\ 50\%} & $61.8\% \pm 1.9\%$ & & \pbox{5cm}{300 \\ 300 \\ 50\%} & $61.8\% \pm 2.4\%$ \\
\noalign{\smallskip}
\hline
\hline
\noalign{\smallskip}
\muacro{WD-GCN} & \pbox{5cm}{\muacro{wd-GC} nodes: \\  \muacro{v-LSTM} nodes: \\ dropout:} &
\pbox{5cm}{$\{100, 200, 300, 400\}$ \\ $\{100, 200, 300, 400\}$ \\ $\{0\%, 10\%, 20\%, 30\%, 40\%, 50\% \}$} & & \pbox{5cm}{300 \\ 300 \\ 50\%} & $70.0\% \pm 3.0\%$ & & \pbox{5cm}{400 \\ 300 \\ 0\%} & $\vect{70.7\% \pm 2.4\%}$ \\
\noalign{\smallskip}
\hdashline[1pt/3pt]
\noalign{\smallskip}
\muacro{CD-GCN} & \pbox{5cm}{\muacro{cd-GC} nodes: \\ \muacro{v-LSTM} nodes: \\ dropout:} &
\pbox{5cm}{$\{100, 200, 300, 400\}$ \\ $\{100, 200, 300, 400\}$ \\ $\{0\%, 10\%, 20\%, 30\%, 40\%, 50\% \}$}  & & \pbox{5cm}{200 \\ 100 \\ 50\%} & $\vect{70.1\% \pm 2.8\%}$ & & \pbox{5cm}{200 \\ 100 \\ 50\%} & $70.5\% \pm 2.7\%$ \\
\noalign{\smallskip}
\hline
\end{tabular}
}
\end{table}

\begin{figure}[t]
{
\hfill
    \begin{subfigure}[t]{0.43\textwidth}
        \centering
        \includegraphics[width=1.15\textwidth]{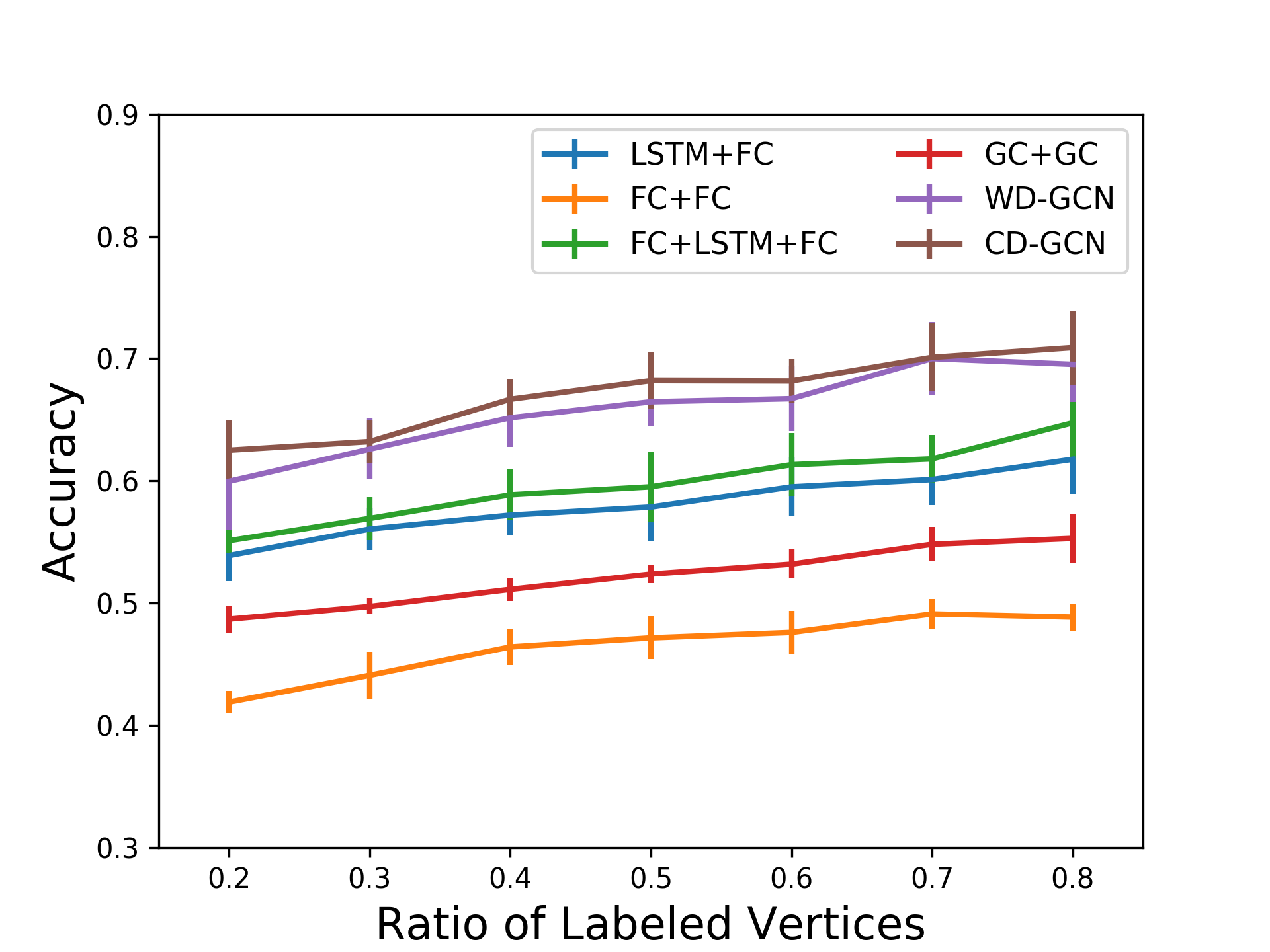}
        \caption{\label{fig:DBLP-results-ratio-labeled-accuracy}Accuracy.}
    \end{subfigure}
    \hfill
        \begin{subfigure}[t]{0.43\textwidth}
        \centering
        \includegraphics[width=1.15\textwidth]{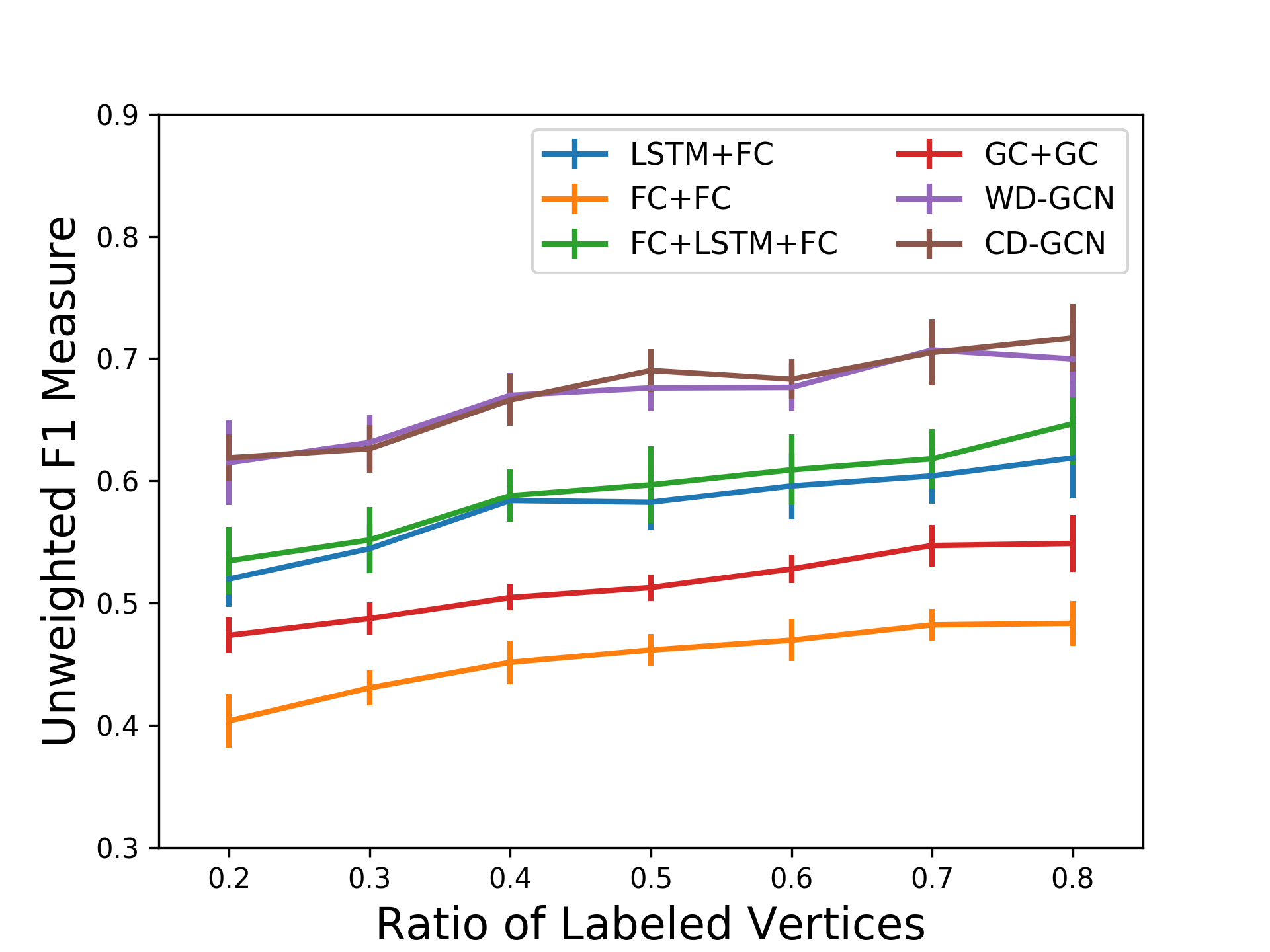}
        \caption{\label{fig:DBLP-results-ratio-labeled-f1}Unweighted F1 Measure.}
    \end{subfigure}
    \hfill}

    \caption{\label{fig:DBLP-results-ratio-labeled}The figure shows the
    performances of the tested approaches (on the \muacro{DBLP} dataset) varying
    the ratio of the labeled vertices. The vertical bars represent the standard
    deviation of the performances achieved in the 10 iterations of the Monte
    Carlo cross-validation.}
\end{figure}

Employing the best configuration for each of the architectures in Table~\ref{tab:DBLP-results-grid}, 
we have further assessed the quality of the tested approaches by evaluating them by changing the ratio of the labeled vertices as follows: $20\%$, $30\%$, $40\%$,
$50\%$, $60\%$, $70\%$, $80\%$. To obtain robust estimations, we have averaged the performances by means of $10$ iterations of Monte Carlo Cross-Validation.
Figure~\ref{fig:DBLP-results-ratio-labeled} reports the results of this experiment.

Both the proposed architectures have obtained promising results that have
overcome those achieved by the considered baselines. Moreover, we have shown
that the \muacro{WD-GCN} and the \muacro{CD-GCN} performances are roughly
equivalent in terms of Accuracy and Unweighted F1 Measure. Architectures such as
\muacro{GCN}s and
\muacro{LSTM}s are mostly likely limited for their inability to exploit jointly
graph structure and long short-term dependencies. Note that, the structure of
the graphs appearing in the sequence is not exclusively conveyed by the DeepWalk
vertex-features, and it is effectively captured by the \muacro{GC} units.
Indeed, the two layers-\muacro{GCN} has obtained better results with respect to
those achieved by the two \muacro{FC} layers.

It is important to underline that, the \muacro{WD-GCN} and the \muacro{CD-GCN}
have achieved better performances with respect to the baselines not for the
reason they exploit a greater number of parameters, or since they are deeper,
rather:
\begin{itemize}[nolistsep]
    \item the baseline architectures have achieved their best performances
    without employing the maximum amount of allowed number of nodes, thus
    showing that their performance is unlikely to become better with an even
    greater number of nodes;
    \item the number of parameters of our approaches is significatly lower than
    the number of parameters of the biggest employed network: \ie the best
    \muacro{WD-GCN} and \muacro{CD-GCN} have, respectively, $872.206$ and
    $163.406$ parameters, while the largest tested network is the
    \muacro{FC}+\muacro{LSTM}+\muacro{FC} with $1.314.006$ parameters;
    \item the \muacro{FC}+\muacro{LSTM}+\muacro{FC} network has a comparable
    depth with respect to our approaches, but it has achieved lower performance.
\end{itemize}

Finally, \muacro{WD-GCN} and \muacro{CD-GCN} have
shown little sensitivity to the labeling ratio, further demonstrating the
robustness of our methods.

\subsubsection{CAD-120}
We have compared the approaches proposed in Section~\ref{sec:supervised-graph-sequence-classification} against a
\muacro{GC}+\muacro{gs-FC} network, a \muacro{vs-FC}+\muacro{gs-FC}
architecture, a \muacro{v-LSTM}+\muacro{gs-FC} network, and a deeper architecture made of \muacro{vs-FC}+\muacro{v-LSTM}+\muacro{gs-FC}. Notice that, for this architectures, the \muacro{vs-FC}s are used with a $\relu$ activation, instead of a $\softmax$.

The $10\%$ of the videos has been selected for testing the performances of the
model, and $10\%$ of the remaining videos has been employed for validation. 

\begin{table}[t]
    \caption{\label{tab:CAD-120-results-grid}Results of the evaluated
    architectures on supervised classification of sequence of graphs employing the
    \muacro{CAD-120} dataset. \muacro{CD-GCN} is the only technique comparing
    favourably to all the baselines, resulting in a Wilcoxon test with a p-value
    lower than $5\%$ for the Unweighted F1 Measure and lower than $10\%$ for the Accuracy.
}
\centering

\resizebox{.9\textwidth}{!}{
\begin{tabular}{lllllcllc}
\cline{5-6}
\cline{8-9}
\noalign{\smallskip}
& & & & \multicolumn{2}{c}{Accuracy} & & \multicolumn{2}{c}{Unweighted F1 Measure} \\
\hline\noalign{\smallskip}
Network & Hyper-params & Grid & \hspace{1em} & \pbox{5cm}{Best \\ Config.} & \pbox{5cm}{Performance\\mean $\pm$ std} & \hspace{1em} & \pbox{5cm}{Best \\ Config.} & \pbox{5cm}{Performance\\mean $\pm$ std} \\
\noalign{\smallskip}
\hline
\noalign{\smallskip}
\muacro{vs-FC}+\muacro{gs-FC} & \pbox{5cm}{\nth{1} \muacro{vs-FC} nodes: \\ dropout:} & 
\pbox{5cm}{$\{ 100, 200, 250, 300 \}$ \\ $\{0\%, 20\%, 30\%, 50\% \}$} & & \pbox{5cm}{100 \\ 20\%} & $49.9\% \pm 5.2\%$ & & \pbox{5cm}{200 \\ 20\%} & $48.1\% \pm 7.2\%$ \\
\noalign{\smallskip}
\hdashline[1pt/3pt]
\noalign{\smallskip}
\muacro{GC}+\muacro{gs-FC} & \pbox{5cm}{\nth{1} \muacro{GC} nodes: \\ dropout:} &
\pbox{5cm}{$\{ 100, 200, 250, 300 \}$ \\ $\{0\%, 20\%, 30\%, 50\% \}$} & & \pbox{5cm}{250 \\ 30\%} & $46.2\% \pm 3.0\%$ & & \pbox{5cm}{250 \\ 50\%} & $36.7\% \pm 7.9\%$ \\
\noalign{\smallskip}
\hdashline[1pt/3pt]
\noalign{\smallskip}
\muacro{v-LSTM}+\muacro{gs-FC} & \pbox{5cm}{\muacro{LSTM} nodes: \\ dropout:} & 
\pbox{5cm}{$\{100, 150, 200, 300\}$ \\ $\{0\%, 20\%, 30\%, 50\% \}$}& & \pbox{5cm}{150 \\ 0\%} & $56.8\% \pm 4.1\%$ & & \pbox{5cm}{150 \\ 0\%} & $53.0\% \pm 9.9\%$ \\
\noalign{\smallskip}
\hdashline[1pt/3pt]
\noalign{\smallskip}
\muacro{vs-FC}+\muacro{v-LSTM}+\muacro{gs-FC} & \pbox{5cm}{\muacro{vs-FC} nodes: \\  \muacro{v-LSTM} nodes: \\ dropout:} &
\pbox{5cm}{$\{100, 200, 250, 300\}$ \\ $\{100, 150, 200, 300\}$ \\ $\{0\%, 20\%, 30\%, 50\% \}$} & & \pbox{5cm}{200 \\ 150 \\ 20\%} & $58.7\% \pm 1.5\%$ & & \pbox{5cm}{200 \\ 150 \\ 20\%} & $57.5\% \pm 2.9\%$ \\
\noalign{\smallskip}
\hline
\hline
\noalign{\smallskip}
\muacro{WD-GCN} & \pbox{5cm}{\muacro{wd-GC} nodes: \\  \muacro{v-LSTM} nodes: \\ dropout:} &
\pbox{5cm}{$\{100, 200, 250, 300\}$ \\ $\{100, 150, 200, 300\}$ \\ $\{0\%, 20\%, 30\%, 50\% \}$} & & \pbox{5cm}{250 \\ 150 \\ 30\%} & $54.3\% \pm 2.6\%$ & & \pbox{5cm}{250 \\ 150 \\ 30\%} & $50.6\% \pm 6.3\%$ \\
\noalign{\smallskip}
\hdashline[1pt/3pt]
\noalign{\smallskip}
\muacro{CD-GCN} & \pbox{5cm}{\muacro{cd-GC} nodes: \\  \muacro{v-LSTM} nodes: \\ dropout:} &
\pbox{5cm}{$\{100, 200, 250, 300\}$ \\ $\{100, 150, 200, 300\}$ \\ $\{0\%, 20\%, 30\%, 50\% \}$} & & \pbox{5cm}{250 \\ 150 \\ 30\%} & $\vect{60.7\% \pm 8.6\%}$ & & \pbox{5cm}{250 \\ 150 \\ 30\%} & $\vect{61.0\% \pm 5.3\%}$ \\
\noalign{\smallskip}
\hline
\end{tabular}
}
\end{table}

Table~\ref{tab:CAD-120-results-grid} shows the results of this experiment. The
obtained results have shown that only \muacro{CD-GCN} has outperformed the
baseline, while
\muacro{WD-GCN} has reached performances similar to those obtained by the
baseline architectures. This difference may be due to the low number of vertices
in the sequence of graphs. Under this setting, the predictive power of the graph
convolutional features is less effective, and the \muacro{CD-GCN} approach,
which augments the plain vertex-features with the graph convolutional ones,
provides an advantage. Hence, we can further suppose that, while \muacro{WD-GCN}
and \muacro{CD-GCN} are suitable to effectively exploit structure in graphs with
high vertex-cardinality, only the latter can deal with dataset with limited
amount of nodes. It is worth noting that, despite all the experiments have shown
a high variance in their performances, the Wilcoxon test has shown that
\muacro{CD-GCN} is statistically better than the baselines with a p-value $<5\%$
for the Unweighted F1 Measure and $<10\%$ for the Accuracy. This reveals that in
almost every iteration of the Monte Carlo Cross-Validation, the \muacro{CD-GCN}
has performed better than the baselines.

Finally, the same considerations presented for the \muacro{DBLP} dataset
regarding the depth and the number of parameters are valid also for this set of
data.

\section{Conclusions and Future Works}\label{sec:conclusion}
We have introduced for the first time, two neural network approaches that are
able to deal with semi-supervised classification of sequence of vertices and
supervised classification of sequence of graphs. Our models are based on
modified \muacro{GC} layers connected with a modified version of \muacro{LSTM}.
We have assessed their performances on two datasets against some baselines,
showing the superiority of both of them for semi-supervised classification of
sequence of vertices, and the superiority of \muacro{CD-GCN} for supervised
classification of sequence of graphs.

We can hypothesize that the differences between the \muacro{WD-GCN} and the
\muacro{CD-GCN} performances when the graph size is small are due to the feature
augmentation approach employed by \muacro{CD-GCN}. This conjecture should be
addressed in future works.

In our opinion, interesting extensions of our work may consist in:
\begin{enumerate*}[label=(\roman*)]
    \item the usage of alternative recurrent units to replace \muacro{LSTM};
    \item to propose further extensions of the \muacro{GC} unit;
    \item to explore the performance of deeper architectures that combine the
    layers proposed in this work.
\end{enumerate*}

\bibliography{ecml_2017}
\bibliographystyle{splncs03}

\end{document}